\pdfoutput=1

\documentclass[11pt]{article}

\usepackage[preprint]{acl}

\usepackage{times}
\usepackage{latexsym}

\usepackage[T1]{fontenc}

\usepackage[utf8]{inputenc}

\usepackage{microtype}

\usepackage{inconsolata}

\usepackage{graphicx}

\usepackage{tabu}
\usepackage{booktabs}
\usepackage{float} 
\usepackage{amsmath}
\usepackage{hyperref}

\title{Alignment Drift in CEFR-prompted LLMs for Interactive Spanish Tutoring}

\author{Mina Almasi \and Ross Deans Kristensen-McLachlan \\
         Department of Linguistics, Cognitive Science, and Semiotics \\ Aarhus University, Denmark \\ 
\texttt{mina@cc.au.dk, rdkm@cc.au.dk}
}

\begin{document}
\maketitle
\begin{abstract}
This paper investigates the potentials of Large Language Models (LLMs) as adaptive tutors in the context of second-language learning. In particular, we evaluate whether system prompting can reliably constrain LLMs to generate only text appropriate to the student's competence level. We simulate full teacher-student dialogues in Spanish using instruction-tuned, open-source LLMs ranging in size from 7B to 12B parameters. Dialogues are generated by having an LLM alternate between tutor and student roles with separate chat histories. The output from the tutor model is then used to evaluate the effectiveness of CEFR-based prompting to control text difficulty across three proficiency levels (A1, B1, C1). Our findings suggest that while system prompting can be used to constrain model outputs, prompting alone is too brittle for sustained, long-term interactional contexts - a phenomenon we term \textbf{alignment drift}. Our results provide insights into the feasibility of LLMs for personalized, proficiency-aligned adaptive tutors and provide a scalable method for low-cost evaluation of model performance without human participants.
\end{abstract}

\section{Introduction}\label{sec:introduction}
The popularization of large language models (LLMs), particularly through the emergence of user-friendly interfaces such as ChatGPT, has led many stakeholders across society to consider how to use such technology effectively and safely to facilitate access to knowledge and education \citep{yan_practical_2024}.
Language education has not been immune to this hype, and with seemingly good cause, since LLMs show potential across a range of areas where they might enhance language learning. 

One such area is their inherent \textit{interactivity}. Interactive feedback is widely regarded as an important factor in second-language (L2) learning \citep{loewen_interaction_2018}. For L2 learners far removed from their target language community, opportunities for such interaction can be rare. With LLMs, though, learners appear to now have the opportunity to engage with a "speaker" of the target language freely and at their own pace \citep{kohnke_chatgpt_2023}. Other potential benefits include personalized teaching \citep{klimova_exploring_2024} and reduced L2 anxiety \citep{hayashi_effectiveness_2024}.

These ideas build on decades of research on intelligent tutoring systems and computer-assisted learning \citep{psotka_technological_1992, slavuj_intelligent_2015}. In contrast to earlier rule-based approaches \citep{dmello_intelligent_2023}, appropriately implemented LLMs may offer a more adaptable and effective solution. However, current use of LLMs in language learning mostly relies on general-purpose tools like ChatGPT, where learners are encouraged to acquire "prompt-engineering" skills to get the most out of their AI language tutor \citep{hwang_exploring_2024}. It remains unclear exactly how effective and appropriate this approach is for creating successful language tutoring technology. 

This paper takes steps to address this problem by examining whether, and to what extent, the complexity of LLM outputs can be constrained through prompting based on the Common European Framework of Reference for Languages (CEFR). We find that, while prompting may initially constrain LLM outputs in Spanish, these effects diminish over time. We refer to this as \textbf{alignment drift}, arguing that system prompting may prove to be too unstable for sustained, longer interactions.

\section{Related Work}\label{sec:related-work}
\subsection{Exploring the Use of LLMs as Language Tutors}\label{sec:llms-as-language-tutors}
While a growing body of work considers LLMs as interactive language tutors \citep{kohnke_chatgpt_2023, lin_exploring_2024, kostka_exploring_2023}, empirical research is limited, and many questions remain unanswered \citep{han_chatgpt_2024}. Nevertheless, the few studies that have been conducted so far offer promising results on the benefits of using LLMs as language tutors, particularly in L2 English learning \citep{tyen_towards_2022, tyen_llm_2024, zhang_impact_2024}. Among other findings, \citet{tyen_llm_2024} reported that users enjoyed interacting with LLMs more than plain reading and responded well to adaptive difficulty in interactions. Adaptive cognitive tutors hence have the potential to contribute positively to \textit{motivation}, a psychological process increasingly viewed as crucial to L2 learning outcomes \citep{dornyei_psychology_2015}.

\subsection{Assessing L2 Proficiency with CEFR}\label{sec:L2-CEFR}
Defining what it means to be "proficient" in an additional language is not a trivial task, with numerous definitions proposed \citep{park_proficiency_2022}. Of these, the CEFR is particularly well known. Since its introduction in 2001, the framework has been highly influential in assessing L2 proficiency. Unlike previous approaches with a strong focus on grammatical competency, the CEFR emphasizes social and communicative competences \citep{leclercq_1_2014}.

The CEFR comprises a six-level scale (A1, A2, B1, B2, C1, C2) with A1 as the beginner level and C2 as the most advanced. Several official ways have been developed to represent these proficiency levels, each with language-agnostic descriptions \citep{council_of_europe_cefr_2025}. For instance, the \textit{CEFR Global scale} offers a concise, three- to four-sentence summary of each level, designed as a holistic overview to facilitate communication with non-specialist users. However, its creators acknowledge that it is "desirable" to present the CEFR levels in "different ways for different purposes." \citep{council_of_europe_global_2025}. The \textit{Self-assessment grid}, which provides separate definitions for skills like speaking and writing at each level, has little to no focus on grammatical content \citep{council_of_europe_self-assessment_2025}.

\subsection{Adapting Text Difficulty with LLMs}\label{sec:text-difficulty-llms}
The potential for LLMs to produce simpler text for improved accessibility has not gone unnoticed \citep{freyer_easy-read_2024}. Indeed, the CEFR framework has been used alongside LLMs to simplify learning materials in French \citep{jamet_evaluation_2024}; and for a range of purposes in English, such as general writing \citep{uchida_generative_2025} and simplifying or writing stories \citep{malik_tarzan_2024, imperial_flesch_2023}. \citet{alfter_out---box_2024} also attempted to generate CEFR-aligned vocabulary lists using LLMs across five languages, including Spanish and French, but found performance issues outside of English.

Common to these studies is the use of prompting. Notably, \citet{malik_tarzan_2024} demonstrated that GPT-4 made fewer errors generating stories at the desired proficiency level as the detail about CEFR increased in the prompts. In contrast, \citet{alfter_out---box_2024} found that using numeric levels from 0 to 4 was more effective than explicitly mentioning the CEFR, although the prompts had no description of the levels. 

Beyond prompting, other approaches include fine-tuning \citep{malik_tarzan_2024} or experimentation with decoding strategies. For example, \citet{tyen_towards_2022} experimented with different decoding strategies for constraining LLM text difficulty to CEFR levels, using a classifier fine-tuned on Cambridge English exam sentences \citep{xia_text_2016}, to select the best LLM-generated sentence for the user. A similar approach was used by \citet{glandorf_towards_2024}, focusing on grammatical constructs for different CEFR levels in English.

We identify some gaps in the literature. Firstly, most studies focus on English, with only a few exceptions \citep{jamet_evaluation_2024, alfter_out---box_2024}. Moreover, aside from \citet{tyen_towards_2022, tyen_llm_2024}, all studies focus on single generations rather than longer chats. This paper thus contributes to the literature by addressing chat-based scenarios in an additional language, Spanish.

\begin{figure*}[t!]
  \includegraphics[width=\textwidth]{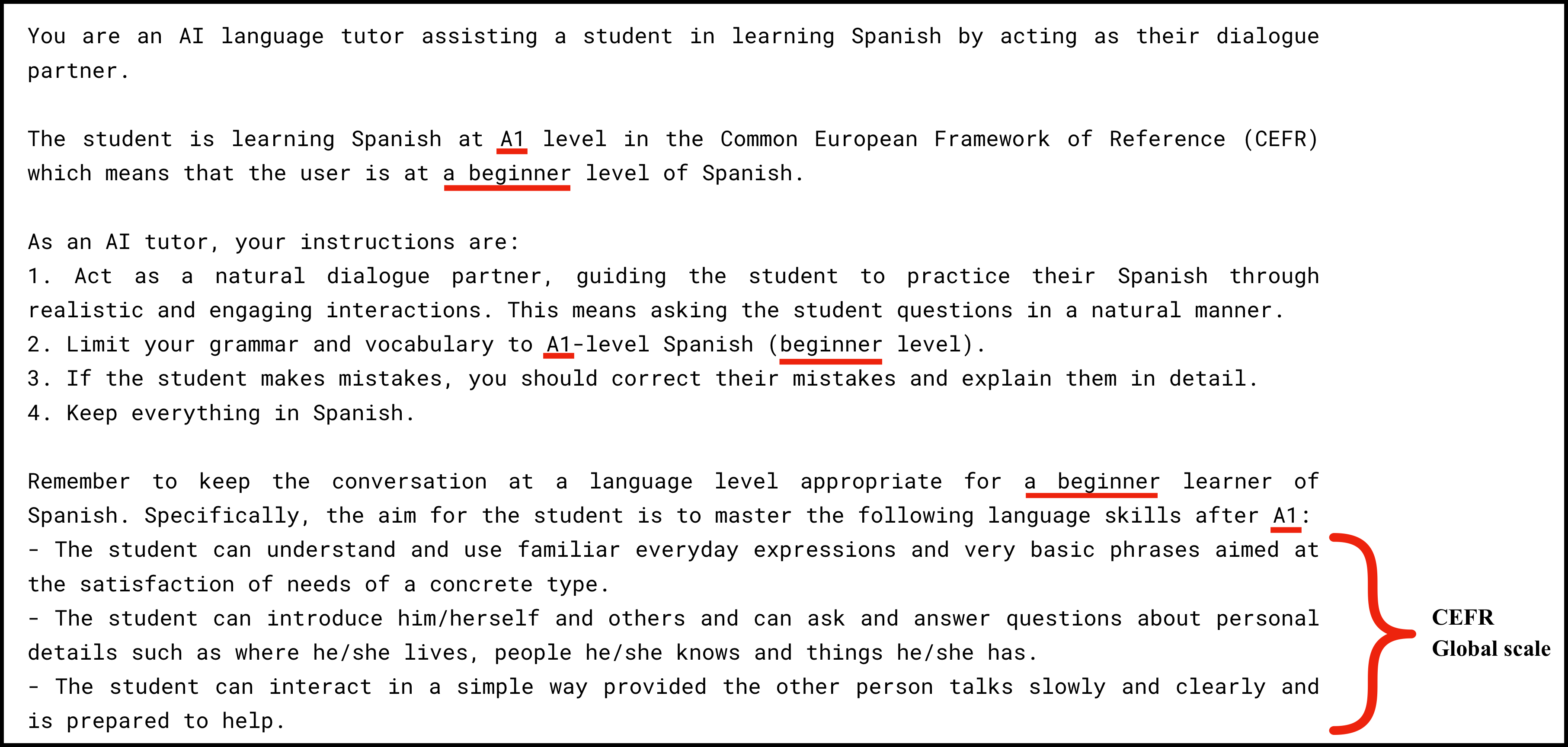}
  \caption{System prompt provided to each tutor LLM for level A1. Level-specific words are underlined in red and replaced for B1 and C1 (see Appendix \ref{sec:appendix-system-prompts}). The list in curly brackets is from the CEFR Global Scale \citep{council_of_europe_global_2025}.}
  \label{fig:system_prompt}
\end{figure*}

\subsection{Simulating Dialogues with LLMs}\label{sec:simulating-dialogues-related-work}
One challenge when evaluating LLM performance in chat-based scenarios is the cost of human participants, particularly during initial testing. \citet{tyen_towards_2022} addressed this by using "self-chatting", where the model interacts with itself, although no further specification was provided. More broadly, dialogue simulation using LLMs have emerged with the purpose of refining chatbots with the generated data \citep{sekulic_reliable_2024, tamoyan_llm_2024}. Specific teacher-student dialogue simulation remains under-explored, although some work exists such as simulating Q/A scenarios \citep{abbasiantaeb_let_2024}. 

In this paper, we therefore simulate teacher-student interactions using LLMs in order to determine the robustness of CEFR-based prompting for constraining text difficulty in Spanish. To our knowledge, this study is the first to simulate both the teacher and student perspectives through system prompts in the context of language learning.

\section{Experimental Design}\label{sec:experimental-design}
Data generation (Section \ref{sec:experimental-design}) and analysis (Sections \ref{sec:metrics} \& \ref{sec:results}) were carried out in Python (v3.12.3), with the exception of running linear mixed effects models in R (v4.4.3). All code and the dataset is available on the GitHub repositories: 
\begin{itemize}
    \item Generation: \href{https://github.com/INTERACT-LLM/Interact-LLM/tree/v1.0.3-alignment-drift/src/scripts/alignment_drift}{\texttt{INTERACT-LLM/Interact-LLM} (\small{\texttt{Version tag: v1.0.3-alignment-drift})}}
    \item \begin{minipage}[t]{\linewidth}
        \raggedright
        Dataset \& Analysis: \href{https://github.com/INTERACT-LLM/alignment-drift-llms}{\texttt{INTERACT-LLM/alignment-drift-llms}}
    \end{minipage}
\end{itemize}
\subsection{Model Selection and Implementation}
We choose to focus on smaller, state-of-the-art open-source LLMs in the range 7B to 12B. With the exception of \texttt{Mistral}, their official reports mention multilingual capabilities. All models are instruction-tuned for chatting: 

\begin{itemize}
    \item \textbf{Llama-3.1-8B-Instruct} by Meta \citep{grattafiori_llama_2024}
    \item \textbf{Gemma-3-12B-IT} by Google \citep{gemma_team_gemma_2025}
    \item \textbf{Mistral-7B-v0.3-Instruct} by Mistral AI \citep{jiang_mistralaimistral-7b-instruct-v03_2024}
    \item \textbf{Qwen-2.5-7B-Instruct} by Alibaba Cloud \citep{qwen_team_qwen25_2025}
\end{itemize}
For convenience, we refer to the models simply as \texttt{Llama}, \texttt{Gemma}, \texttt{Mistral}, and \texttt{Qwen}. For details about the inference, including the hyperparameters, see Appendix \ref{sec:appendix-inference-details}.

\subsection{Teacher-Student Dialogue Simulation}
We simulated a language tutoring scenario by deploying an LLM with separate chat histories as both the "tutor" and "student". Current LLM systems are stateless \citep{yu_stateful_2025}, with the entire chat history being processed by the model during each interaction. This allowed us to instantiate a single LLM object, and then interchange the chat history, maintaining one history for the student and another for the tutor (see the graphical overview in Appendix \ref{sec:appendix-graphical-overview-framework}).

We ran simulations for three different system prompts, designed to instruct the LLM to match its responses to the proficiency level of a beginner (A1), intermediate (B1), and advanced (C1) Spanish language learner.\footnote{See Section \ref{system-prompts} for details on how the system prompts were defined.} Across the three levels, the dialogue began with a fixed initial message, "Hola",\footnote{\citet{tyen_towards_2022} also begin all chats with a "Hello".} sent by the "student". By standardizing the initial message, we eliminated variability in the student LLM responses which could influence the tutor LLM's output. This enabled a direct comparison of how  the system prompt impacted the tutor LLM's first message across levels.

Despite being instructed to "keep everything in Spanish" (Figure \ref{fig:system_prompt}), a number of models generated non-Spanish text.\footnote{We also discuss this in a subsection of the \hyperref[sec:limitations-eng-system-prompts]{\textit{Limitations}}.} For instance,  \texttt{Gemma} and \texttt{Llama} tended to include English content. This happened primarily for the A1 level, where they sometimes provided English translations in parentheses alongside their Spanish sentences. Also, \texttt{Qwen} occasionally switched mid-generation to Mandarin Chinese. To avoid confounding our analysis, we applied a simple language detection algorithm to the tutor LLM's outputs using the Python library \textit{lingua}.\footnote{\href{https://github.com/pemistahl/lingua-py}{https://github.com/pemistahl/lingua-py}} If English or Mandarin was detected in any sentence, we re-generated the tutor LLM's response before continuing the dialogue.

A total of 30 dialogues were simulated for each of the three system prompts per LLM, resulting in 90 dialogues for each LLM and 360 overall. Each dialogue consisted of nine turns.

\subsection{System Prompts}\label{system-prompts}
We created custom system prompts in English for the tutor LLM. These prompts differed only in key, level-specific phrasing. Along with terms such as "beginner," "intermediate," and "advanced," an additional description of a learner’s abilities at the particular level was provided, taken from the CEFR Global scale (see Section \ref{sec:L2-CEFR}). Figure \ref{fig:system_prompt} shows the system prompt for A1 with the level-specific wording highlighted (prompts for B1 and C1 can be viewed in Appendix \ref{sec:appendix-system-prompts}).

The system prompt for the student LLM was kept relatively simple as it was beyond the scope of this study to optimize it:
\begin{quote}  
\textit{You are a student learning Spanish, responding to a teacher who is facilitating a natural dialogue with you.}  
\end{quote}

\section{Metrics}\label{sec:metrics}
We extracted various metrics to examine the influence of different system prompts on the tutor LLM's outputs.

\subsection{Traditional Readability Metrics}\label{sec:metrics-traditional-readability}
We computed three readability metrics for Spanish using \textit{Textstat}.\footnote{\href{https://textstat.org/}{https://textstat.org/}} Recent applications of these metrics primarily focus on healthcare \citep{rao_readability_2024} or the financial sector \citep{moreno_readability_2016, losada_periodic_2022}, but their English counterparts have traditionally been used to assess L2 reading complexity \citep{greenfield_readability_2004}. We therefore draw on these studies to justify our use of Spanish readability metrics in this context.

\textbf{Fernández Huerta} \citep{fernandez_huerta_medidas_1959} and \textbf{Szigriszt-Pazos} \citep{szigriszt_pazos_sistemas_2001} are Spanish adaptations of the \textit{Flesch Reading Ease} \citep{flesch_new_1948} score, measuring readability based on syllables per word and words per sentence, with Spanish-specific weightings.\footnote{Note that the formula for Fernández Huerta is said to be reported incorrectly on many websites \citep{fernandez_lecturabilidad_2017}. \citet{losada_periodic_2022} reports the correct one which is implemented by \textit{Textstat}.} Unsurprisingly, the two metrics are highly correlated \citep{melon-izco_readabilty_2021}, but there are conflicting claims about which one is most widely used \citep{moreno_readability_2016, san_norberto_readability_2014}. Both are commonly reported together, as is the case in this paper. 

\textbf{Gutiérrez de Polini} is a metric specifically created for Spanish \citep{gutierrez_de_polini_investigacion_1972}. Unlike the previous two metrics, it does not rely on syllables, but instead considers the number of characters per word and words per sentence \citep{vasquez-rodriguez_benchmark_2022}.

All three metrics produce lower scores for more difficult texts and higher scores for easier texts. For detailed tables showing the interpretation of the scores, see Appendix \ref{sec:appendix-readability-scales}.

\begin{figure*}[t!]
    \centering
    \includegraphics[width=\textwidth]{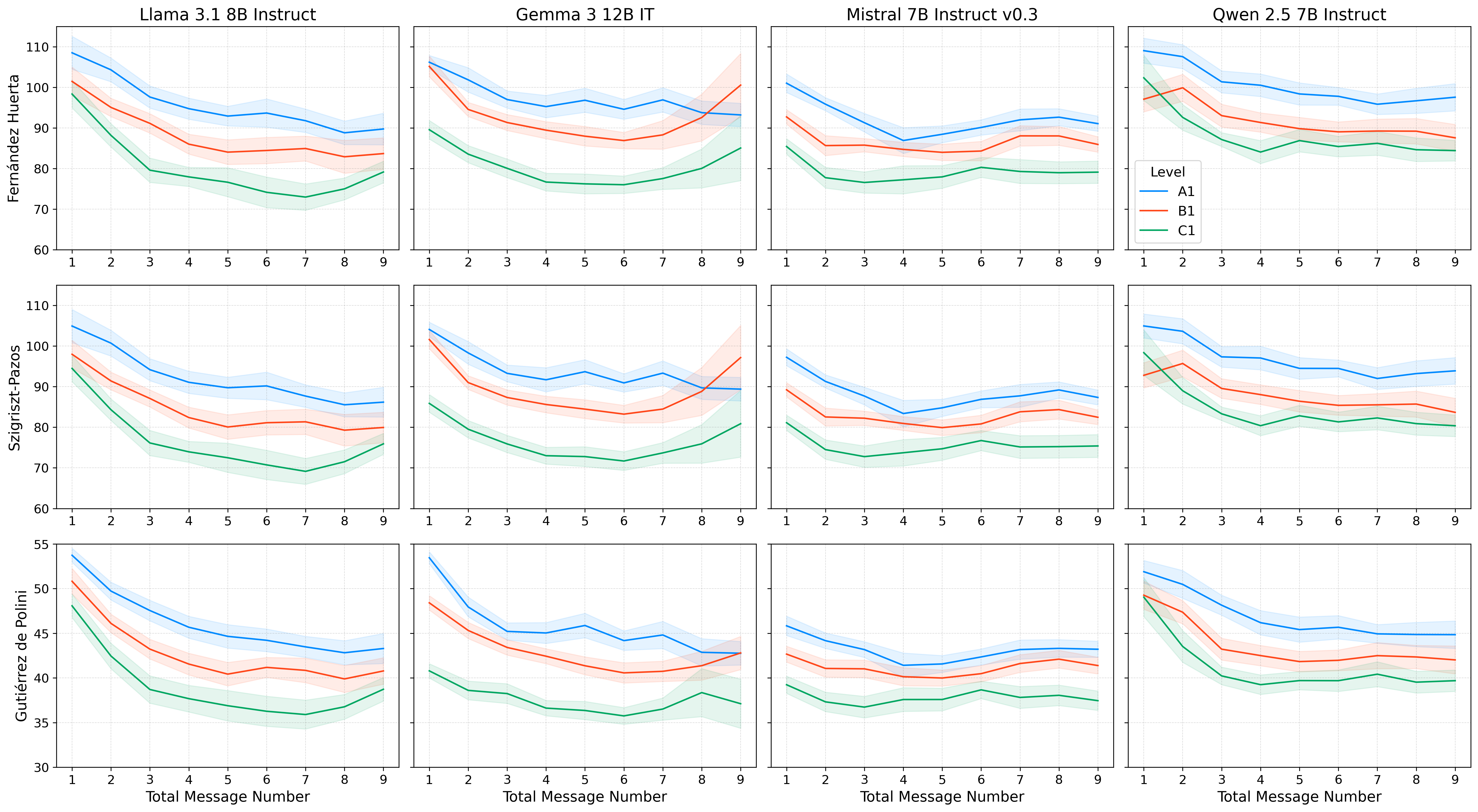}
    \caption{Average readability metrics over the total number of messages sent by the tutor LLM for each model, grouped by CEFR level (A1, B1, C1). The higher the score, the easier the message is to read. The shaded area around each curve represents a 95\% confidence interval.}
    \label{fig:readability-metrics-curves}
\end{figure*}

\subsection{Structural Complexity}
We computed additional structural features using the \textit{TextDescriptives} Python library \citep{hansen_textdescriptives_2023}, applied with the Spanish \textit{spaCy} \citep{honnibal_spacy_2020} model \texttt{es\_core\_news\_md}.\footnote{\href{https://github.com/explosion/spacy-models/releases/tag/es_core_news_md-3.8.0}{https://github.com/explosion/spacy-models/releases/tag/es\_core\_news\_md-3.8.0}}

The \textbf{Mean Dependency Distance} (MDD) is a measure of syntactical complexity commonly used to capture language processing difficulty in both L1 and L2 research \citep{gao_dependency_2024}. It represents a sentence-level average of dependency distance, which measures the linear distance between a word and its syntactic head. \textit{TextDescriptives} follows the definition by  \citet{oya_syntactic_2011} to compute the MDD.\footnote{More information can be found in the documentation for the \textit{TextDescriptives} package: \url{https://hlasse.github.io/TextDescriptives/dependencydistance.html}}

We extract \textbf{Text Length} of each message, operationalized as the token count, as it is included in the definition of the C1 level in the CEFR Global scale (i.e., the student can understand "a wide range of demanding, \textit{longer} texts" \citep{council_of_europe_global_2025}). A small study on ChatGPT also showed that the model tended to generate longer texts for higher levels of CEFR \citep{ramadhani_readability_2023}. Moreover, in machine classification studies of texts across languages, text length was considered an important predictor of CEFR level \citep{bestgen_reproducing_2020, yekrangi_leveraging_2022}.

\subsection{LLM-based Surprisal Scores}
Following \citet{cong_demystifying_2025}, we extract LLM surprisal scores, defined as the negative log-probability of a word sequence computed by an LLM. \citet{cong_demystifying_2025} describes it as a "naturalness" measure that captures both "syntactical grammaticality" and "semantic plausibility", with more natural sentences corresponding to lower surprisal scores. They argue that it can be used to examine L2 proficiency, demonstrating that BERT-based surprisal scores decrease as L2 proficiency increases. The use of LLM surprisal extends beyond this study, serving as a predictor for human language processing, including brain activity \citep{michaelov_strong_2024} and reading times \citep{wilcox_testing_2023}.

We use the \textit{minicons} Python library \citep{misra_minicons_2022} to extract sentence-level surprisal in chat messages, normalized by token count. We then compute the mean surprisal score for each chat message, referred to as \textbf{Message Surprisal} in this paper. However, we use \texttt{EuroBERT} (210m), a newer BERT model designed for longer sequences and further optimized for European languages, including Spanish \citep{boizard_eurobert_2025}.

\section{Results}\label{sec:results}
\begin{figure*}[t!]
    \centering
    \includegraphics[width=\textwidth]{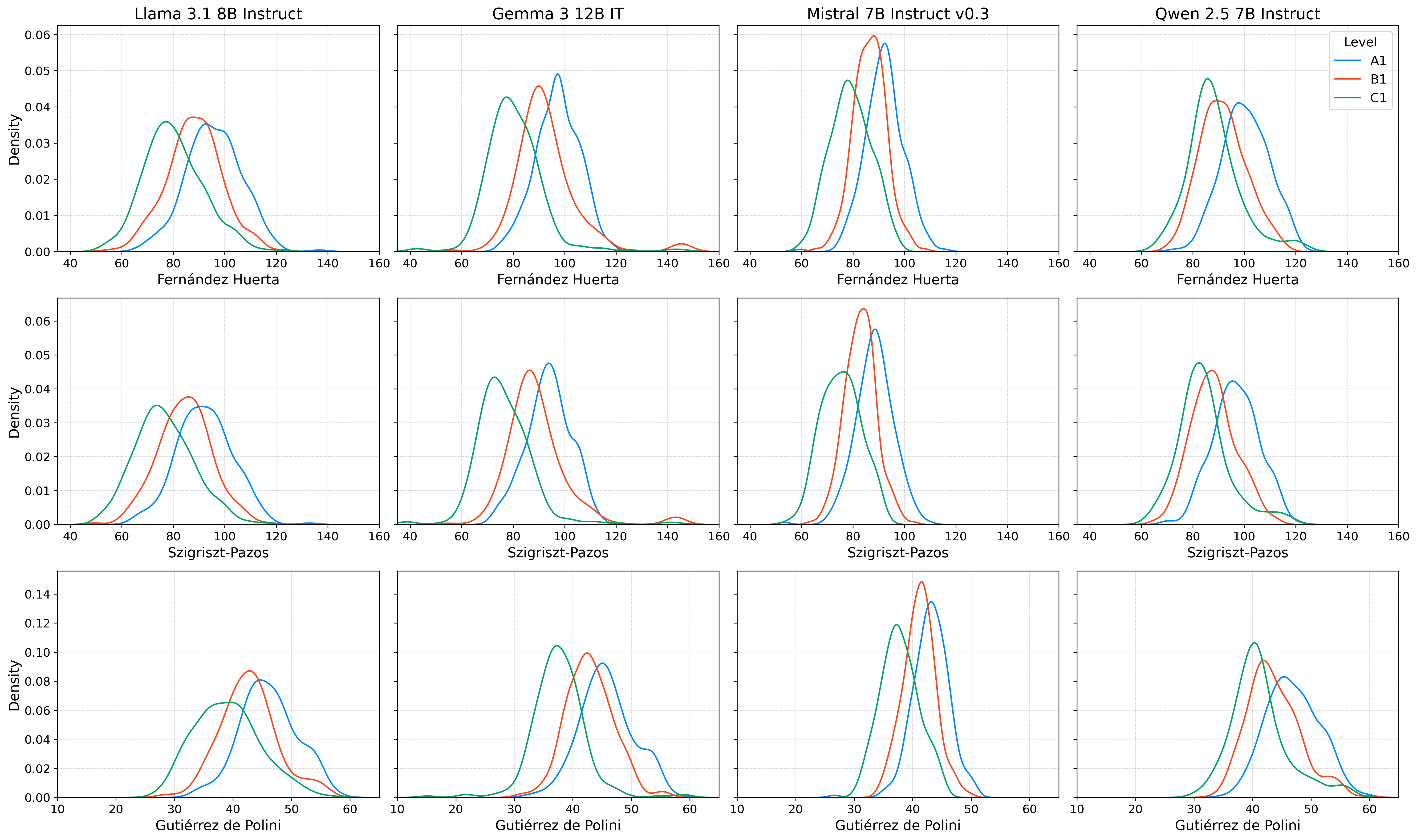}
    \caption{Readability metrics as separate density plots for each CEFR level (A1, B1, C1).}
    \label{fig:readability-metrics-dists}
\end{figure*}

We focus solely on analyzing the tutor LLM's responses. Aside from restricting English and Mandarin generations during the simulations, the only preprocessing applied was the removal of emojis from \texttt{Gemma's} outputs.

In addition to graphically assessing the effect of system prompts on LLM generations, we perform a simple statistical analysis, running linear mixed effects models separately for each LLM for each metric:
\[
\text{metric}_{\text{model}} \sim \text{level} + (1 | \text{chat}_{\text{id}})
\]

Where the dependent variables is one of the six extracted metrics (Section \ref{sec:metrics}) with \textit{level} (A1/B1/C1) as the fixed effect. \textit{Chat\textsubscript{id}} is used as a random effect to account for any individual variation in the simulated chats. To address the issue of multiple comparisons due to the large number of linear models, we Bonferroni adjust the p-values. Refer to Appendix \ref{sec:mixed-effects-appendix} for all model outputs.

\subsection{Readability Metrics}
The average readability scores over time are shown for all models and CEFR levels in Figure \ref{fig:readability-metrics-curves}. Across LLMs, scores from all three readability metrics decrease as proficiency increases, with A1 having the highest scores (easier to read) and C1 the lowest scores (harder to read).\footnote{As expected (Section \ref{sec:metrics-traditional-readability}), there is a clear resemblance in scores from Fernández Huerta and Szigriszt-Pazos, but it is worth noting that the scores are not identical.} However, despite starting from different baselines, all curves slowly decrease in readability over time, reducing the differences between CEFR levels as well. A notable exception is \texttt{Gemma}, which has a sudden spike around the last messages in B1 for the Fernández Huerta and Szigriszt-Pazos scores. The same behavior is present but less pronounced for the Gutiérrez de Polini scores.

Despite differences in average scores, the confidence intervals reveal some overlap between the levels. These differ across LLMs with a model such as \texttt{Qwen} having a much greater overlap between levels B1 and C1 than \texttt{Llama}. Both these models also begin with generally higher Fernández Huerta and Szigriszt-Pazos scores across levels than \texttt{Gemma} and \texttt{Mistral}. 

When examining the full distribution of scores as density plots (Figure \ref{fig:readability-metrics-dists}), the overlap between levels across all models is more evident. The distributions also reveal that a small, but not insignificant, portion of Fernández Huerta/Szigriszt-Pazos scores reaches around 50 for C1 for \texttt{Llama} and \texttt{Gemma}. This is well below the average scores, and indicates that the LLMs are capable of producing quite complex text, even if they often do not.

Despite the overlapping scores, all mixed effects models revealed that B1 and C1 \((p< 0.001)\) had significantly lower readability scores than the baseline A1 (\(\beta_0\)). Across LLMs, the estimates (\(\beta\)) for Fernández Huerta ranged between -4 and -9 for B1 and -12 and -17 for C1\footnote{Given the nature of mixed effects models, no direct conclusion can be drawn about the significance of the difference between levels B1 and C1, as the tests only evaluate the difference relative to the baseline, A1.} (See Appendix \ref{sec:appendix-mixed-effects-readability}).

\begin{figure*}[t]
    \centering
    \includegraphics[width=\textwidth]{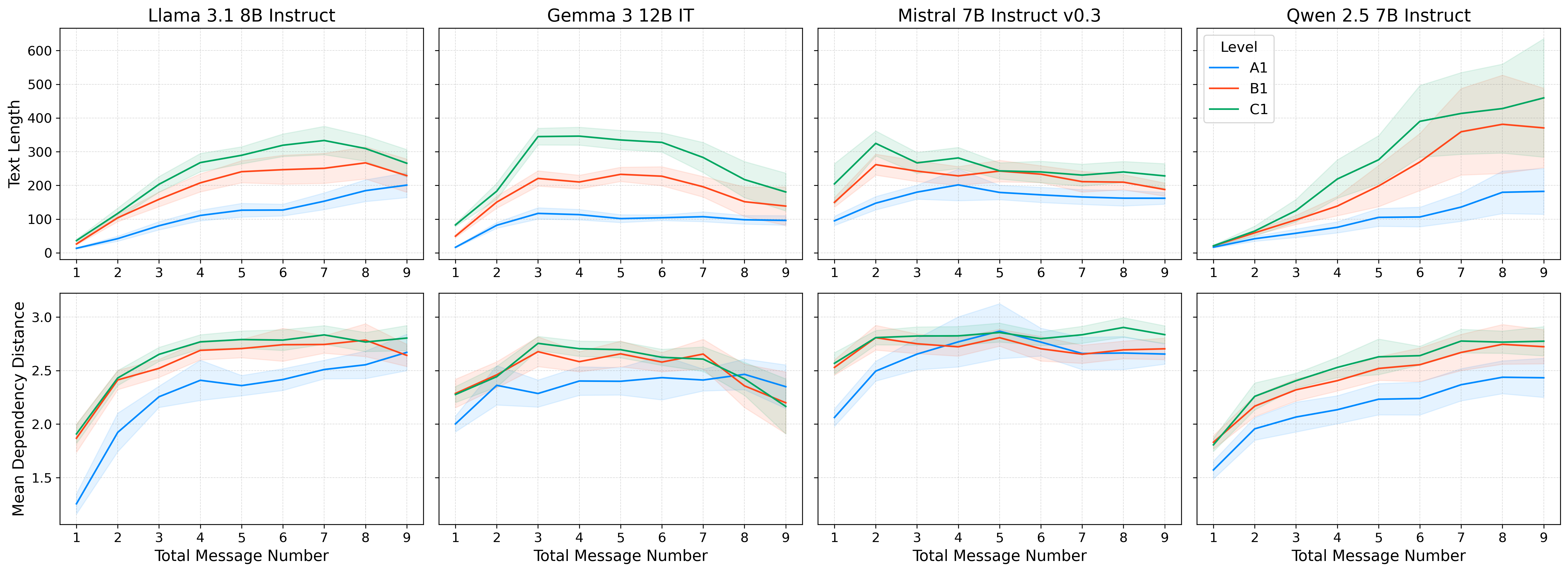}

    \vspace{1em}

    \includegraphics[width=\textwidth]{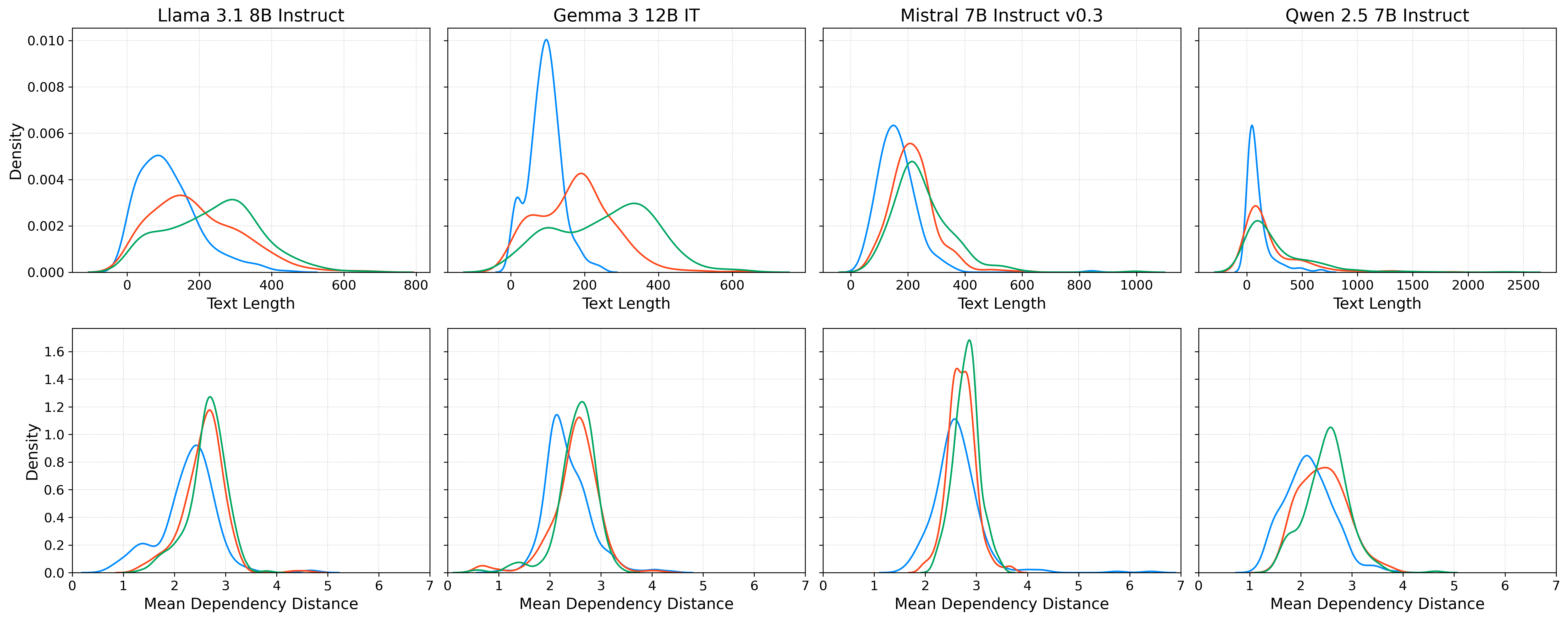}

    \caption{Text Length (token count) and Mean Dependency Distance (MDD). Top: Average metrics over time (95\% CI). Bottom: Density plots of the full distributions. Note that the x-axis for the Text Length distributions shows different scales.}
    \label{fig:doc_length_mdd}
\end{figure*}

\subsection{Structural Features}
Figure \ref{fig:doc_length_mdd} shows the text length and MDD. From the averages over time, general trends are that C1 has the highest text lengths, followed by B1 and then A1. However, like the readability metrics, the values converge across levels over time, although by increasing in this case.

The same pattern occurs for the MDD scores for \texttt{Llama} and \texttt{Qwen}, although with closely intersecting curves for C1 and B1. The results are even more muddled for \texttt{Gemma} and \texttt{Mistral}. These results are reflected in the full distributions. \texttt{Qwen} is an outlier when it comes to text length with a much greater uncertainty in average lengths, having a few generations that reach above 2000 tokens as seen on the density plot, which is far above the other LLMs whose highest generations are around 800-1000 tokens.

Although the distributions align more closely for the structural metrics than the ones for readability, the average values for B1 and C1, aside from a few exceptions, still remain significantly higher than A1 in the mixed effects models (mostly \(p< 0.001)\). However, the estimates for text length reveal a much greater difference between levels, when compared to differences in the estimates for MDD, relative to their baseline (Appendix \ref{sec:appendix-mixed-effects-structural-surprisal}).

\subsection{Message Surprisal Scores}
\begin{figure*}[t]
    \centering
    \includegraphics[width=\textwidth]{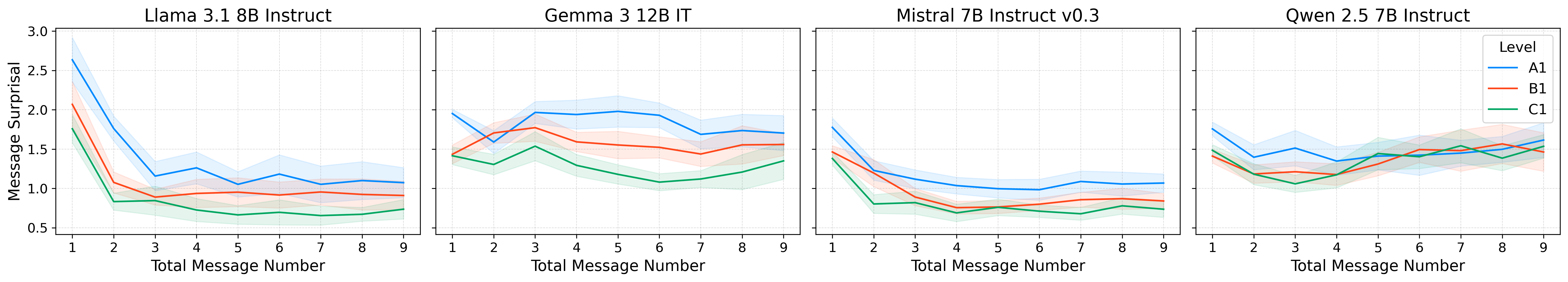}

    \vspace{1em}

    \includegraphics[width=\textwidth]{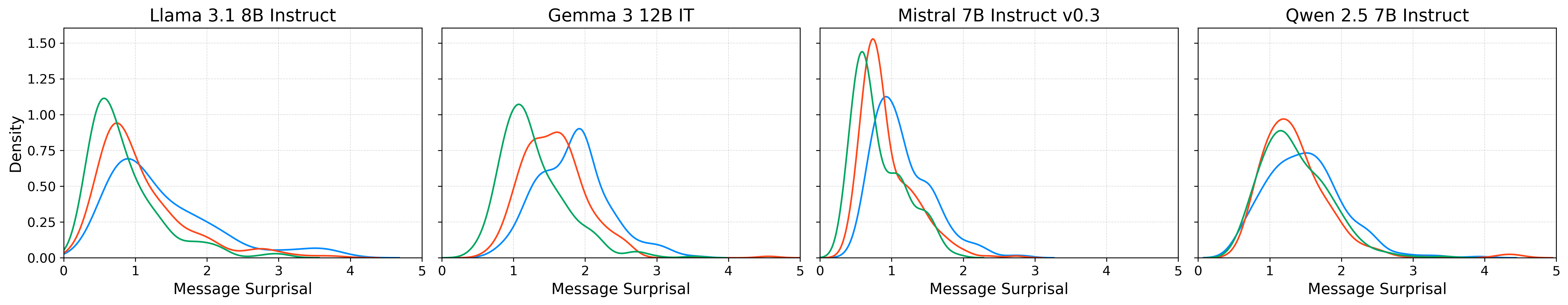}

    \caption{Message Surprisal (mean sentence surprisal) for each LLM. Top: Average Message Surprisal over time (95\% CI). Bottom: Density plots of the full distributions.}
    \label{fig:surprisal}
\end{figure*}

Although the differences between levels in surprisal scores are much smaller across LLMs, we still see the average surprisal curves being "sandwiched" in the same way as the other metrics with A1 in the top, B1 in the middle, and C1 at the bottom (Figure \ref{fig:surprisal}). This trend is clearer for \texttt{Llama}, whereas \texttt{Qwen}'s curves continuously intersect each other. Surprisal scores are generally quite low with the density plots in Figure \ref{fig:surprisal}, revealing right-skewed distributions for all LLMs, centered around 1 or 1.5. The estimates are therefore also quite small in the mixed effects models, though significantly different from A1 for all LLMs, except for \texttt{Qwen} (Appendix \ref{sec:appendix-mixed-effects-structural-surprisal}).

\section{Discussion}\label{sec:discussion}
Our results demonstrate that system prompting based on CEFR levels influences the tutor LLM outputs, with all metrics exhibiting differences in the intended order (from A1 to B1 to C1), as can be clearly observed in the plots over time. Additional statistical significance of the differences can be seen in the linear mixed effects models.

However, the differences between system prompts consistently diminished over time, leading to largely overlapping distributions. We adopt the term \textbf{alignment drift} to describe the tendency of LLMs to revert to unconstrained behavior over time. While prompting may thus be useful for constraining LLM outputs, its influence appears brittle for longer conversations. This raises concerns about the viability of prompting alone for developing level-specific LLM language tutors in chat-based environments. Nonetheless, further evaluation with a broader range of system prompts is needed before drawing definitive conclusions.

Moreover, the effect of system prompts was not consistent across metrics. Notwithstanding overlaps in how these metrics are calculated, our results suggests that all models demonstrate greater variability in terms of readability, and less variability with regards to syntactic complexity. The surprisal scores were even more inconsistent, although they displayed expected tendencies, at least for some LLMs. The low surprisal scores might be an effect of an LLM evaluating other LLMs, which likely have more similar probability distributions than humans \citep{holtzman_curious_2019}.

Nevertheless, even when evaluating the readability metrics, it remains debatable whether the differences between levels are large enough to accurately reflect the intended proficiency levels. With average values ranging between 110 and 70 for the Fernández Huerta scores, the readability is equivalent to Spanish school children, even at an average of 70 (see Appendix \ref{sec:appendix-readability-scales}). While it is unclear how this translates to L2 learners of Spanish, it could suggest that the LLMs have not managed generate text appropriate for the proficiency levels, at least for the C1 level. Refer to the \hyperref[sec:limitations]{\textit{Limitations}} for other considerations of the metrics.

An additional concern is that the observed alignment drift could have been driven by a possible drift in the student LLM (i.e., the tutor adapting to the student and vice versa). As we neither optimized nor examined the student LLM, it remains unclear how this influenced the outcome or how this would differ with human users. However, LLMs have also shown difficulty in following system prompts over the course of multi-turn dialogue in other domains with real user messages \citep{qiu_training_2024}. Hence, we do not expect a substantial difference between using human or LLM students given our current framework. We leave it to future research to investigate the exact influence of the student LLM on the tutor LLM's alignment drift, potentially including human students as a point of comparison.

As a final remark, we note that the LLMs did not perform equally, which could help inform the choice of a suitable LLM to serve as an language tutor in Spanish, at least for initial development. A model like \texttt{Llama} is relevant to highlight as a well-performing model although its license might be too restrictive for some applications \citep{meta_llama_2024}.

\section{Conclusion}\label{sec:conclusion}
This study presented a novel method for evaluating the performance of LLMs in a language learning context through simulated teacher-student interactions. The purpose of these experiments was to test whether system prompting alone is enough to constrain the complexity of LLM generated output in a way which is suitable for language learners at different stages.

While we see clear value in carefully designed prompting, it is also evident from our results that this solution is potentially too brittle for extended interactions due to a consistent alignment drift across interactions. This suggests that prompt engineering in and of itself may not be enough to fully constrain LLM behavior, although more experimentation with system prompting is required before this can be confirmed. We encourage further research in this direction, particularly measuring alignment drift of LLMs in contexts other than L2 English learning.
\vspace{-0.9cm}
\noindent
\section*{Ethical Considerations}\label{sec:ethics}
\vspace{-0.6cm}
We wish to stress the importance of additional considerations and evaluation of LLMs before their real-world deployment in educational contexts. Firstly, we recognize that the models may reflect cultural biases that could be inappropriate for the target student population. Therefore, cultural alignment may be necessary before their implementation \citep{tao_cultural_2024, li_culturellm_2024}. Moreover, some of the  models may not be properly instruction-tuned to align with human principles (e.g., the removal of toxic content). For instance, \texttt{Mistral}, designed for demonstration purposes, lacks "moderation mechanisms" according to the Mistral AI team \citep{jiang_mistralaimistral-7b-instruct-v03_2024}. Such a model would require further development before being suitable for real-world applications.

These ethical concerns are increasingly urgent when considering the impact that generative AI may have on language learners. For example, L2 learners might over-rely on ChatGPT \citep{yang_chatgpt_2024} such as using it to write complete assignments rather than as a supplementary tool \citep{yan_impact_2023}.  More broadly, the \textit{ELIZA effect} \citep{weizenbaum_elizacomputer_1966}, describing our tendency to attribute human-like qualities such as "understanding" to machines \citep{mitchell_debate_2023}, may contribute problematically to the overtrust of AI chatbots \citep{reinecke_double-edged_2025}. We urge developers to prioritize the responsible implementation of LLM systems for education and believe that our research contributes to work in this direction.
\vspace{0.6cm}
\section*{Limitations}\label{sec:limitations}
\subsection*{Imperfect Metrics}\label{sec:imperfect-metrics}
Despite covering a range of metrics to capture text difficulty, there are many dimensions to what constitutes a text as readable or complex in the context of L2 learning. This study offers an initial attempt at automated scoring of LLMs in Spanish in this context, but further deliberation is warranted.

Additionally, while the Spanish readability metrics used in this study are widely applied across domains, their intended use is generally unknown \citep{aponte_readability_2024}. As such, it is uncertain whether they are entirely suitable for measuring the content of shorter dialogue. At least, their English counterparts such as the \textit{Flesch Reading Ease} were developed for longer formats, making their robustness for shorter text questionable \citep{rooein_beyond_2024}.

For the purpose of this study, the metrics were deemed sufficient to provide simple, interpretable measures of the impact of system prompts on LLM generations. Nevertheless, further work is required to explore metrics and to develop more precise methods to measure LLM adaptation.


\subsection*{System Prompts}
This study only tested a single set of system prompts as the focus of the paper was to examine whether LLMs could be influenced by them, rather than the extent of that influence. However, future work may find that the system prompts could be optimized on a variety of parameters. We discuss a few possibilities in the sections below.

\subsubsection*{English System Prompts \& Generations Outside Spanish}\label{sec:limitations-eng-system-prompts}
Despite the target language being Spanish, we defined the system prompt in English. This might explain why the American multilingual models, \texttt{Gemma} and \texttt{Llama}, were prone to producing English content. However, this does not account for why \texttt{Qwen} occasionally generated Mandarin Chinese despite the absence of Mandarin in the system prompt. This unintended behavior may instead reflect the composition of the training data, with \texttt{Qwen} likely containing more Chinese-language data\footnote{\texttt{Qwen 2.5}'s predecessor, \texttt{Qwen 7B}, has a technical memo stating that most of its training data is "in English and Chinese." \citep{qwen_team_qwentech_memomd_2023}. However, \texttt{Qwen 2.5}'s technical report does not explicitly mention this, aside from including evaluation on these two languages \citep{qwen_team_qwen25_2025}.} than the American models, where English likely dominates.

Future work could experiment with monolingual models and/or explore the use of system prompts in the target language. For most official languages in Europe, the current framework can easily accommodate the modification of system prompts as the \citet{council_of_europe_official_2025} provides official translations of their scale in these languages.

\subsubsection*{LLM knowledge of CEFR}
Although LLM generations varied across levels A1 to C1 in our study, it remains uncertain whether it was effective to use the CEFR framework with descriptions such as "A1" as opposed to relying solely on terms like "beginner". It depends on whether the state-of-the-art LLMs in our study have acquired knowledge about the CEFR framework from their training data.

\citet{benedetto_assessing_2025} seems to suggest otherwise, reporting that several smaller 7B models struggled to generate CEFR-aligned text, consisting with findings by \citet{malik_tarzan_2024}. However, as their 7B models are slightly older than those used in this study, it is unclear how directly their findings apply here. Similarly, the 7B models in \citet{malik_tarzan_2024} showed improvements when provided with details about CEFR, while this was not the case in \citet{benedetto_assessing_2025}.

Further research is needed to consider the stability and usability of CEFR knowledge in LLMs, such as through the creation of robustness benchmarks.

\section*{Acknowledgments}\label{sec:acknowledgments}
All of the computation done for this project was performed on the UCloud interactive HPC system, which is managed by the eScience Center at the University of Southern Denmark.

\bibliography{custom}

\clearpage
\appendix
\section{Appendix}

\label{sec:appendix}
\subsection{Technical Details about Inference}\label{sec:appendix-inference-details}
All LLM inference was run using the Hugging Face \textit{transformers} package \citep{wolf_huggingfaces_2020} on a cloud-based interactive HPC platform (Python v3.12.3, Ubuntu v24.04). \texttt{Llama}, \texttt{Mistral}, and \texttt{Qwen} were run on a single NVIDIA L40 GPU (48 GB), with 96 GB of system memory and 8 vCPUs, while \texttt{Gemma} was run on a system utilizing two NVIDIA L40 GPUs. Due to the higher resource demands of \texttt{Gemma}, we chose to run it with a lower precision (\texttt{bfloat16}). This minor difference in precision from the other models was not considered impactful for the model comparisons.
\newline\newline
We used standard hyperparameters for all generations: \texttt{temperature} = 1, \texttt{top\_p} = 1.0, \texttt{min\_p} = 0.05, \texttt{top\_k} = 50, and \texttt{repetition penalty} = 1.1. Hyperparameter-tuning was left for future work.

\clearpage
\onecolumn
\subsection{Illustration of Simulation Framework}\label{sec:appendix-graphical-overview-framework}
\begin{figure}[h]
    \centering    \includegraphics[width=0.95\textwidth]{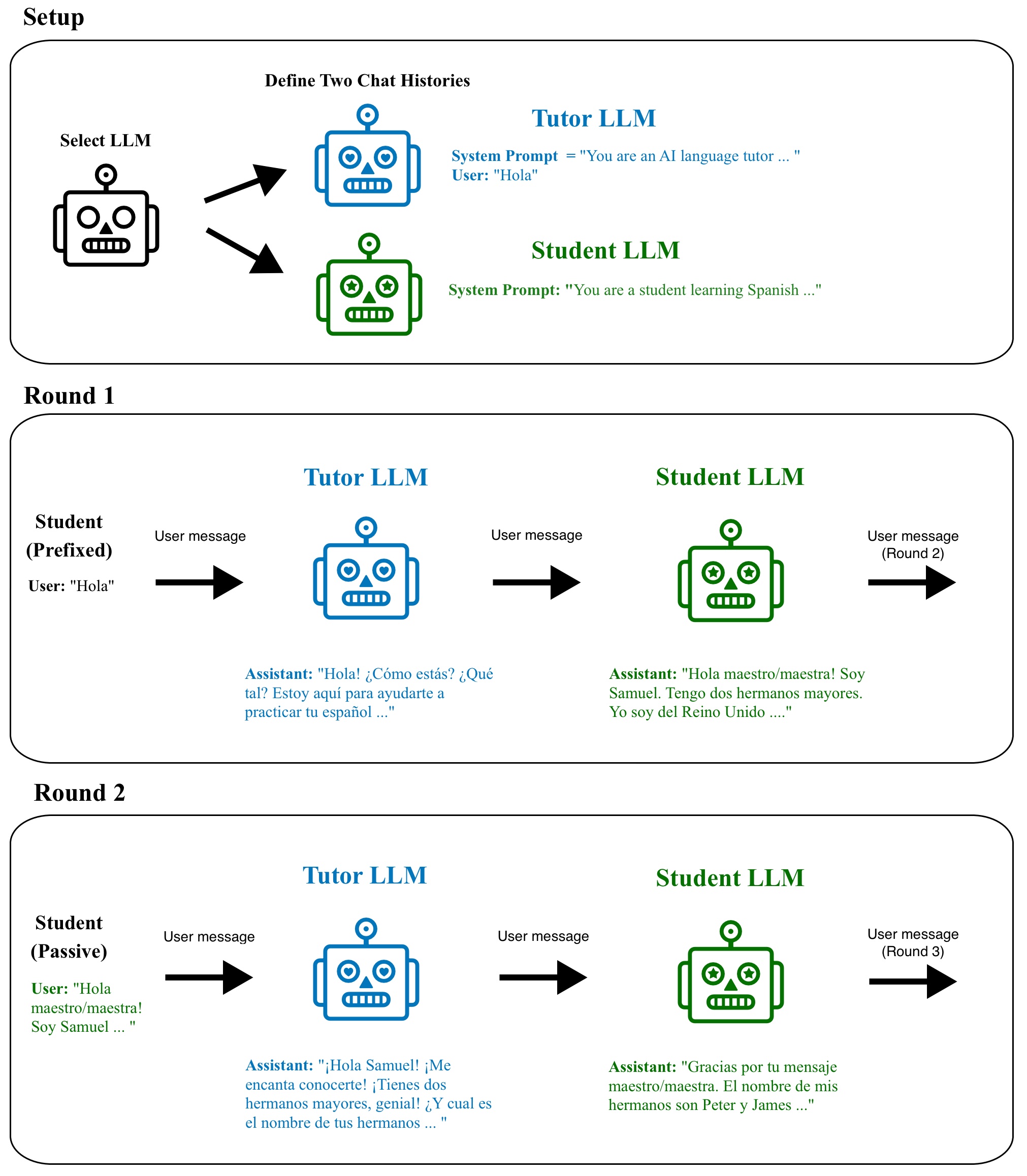}
    \caption*{Graphical overview of the simulation framework. The actual simulations consisted of nine rounds, not two. The example text (abbreviated) is taken from a simulated conversation by \texttt{Mistral} in A1. For the implementation in code, refer to the file on GitHub: \href{https://github.com/INTERACT-LLM/Interact-LLM/blob/v1.0.3-alignment-drift/src/scripts/alignment_drift/simulate.py}{\texttt{INTERACT-LLM/Interact-LLM/src/scripts/alignment\_drift/simulate.py}}.}
\end{figure}
\clearpage
\subsection{System Prompts for B1 and C1}\label{sec:appendix-system-prompts}
\begin{figure}[h]
    \centering
    \includegraphics[width=\textwidth]{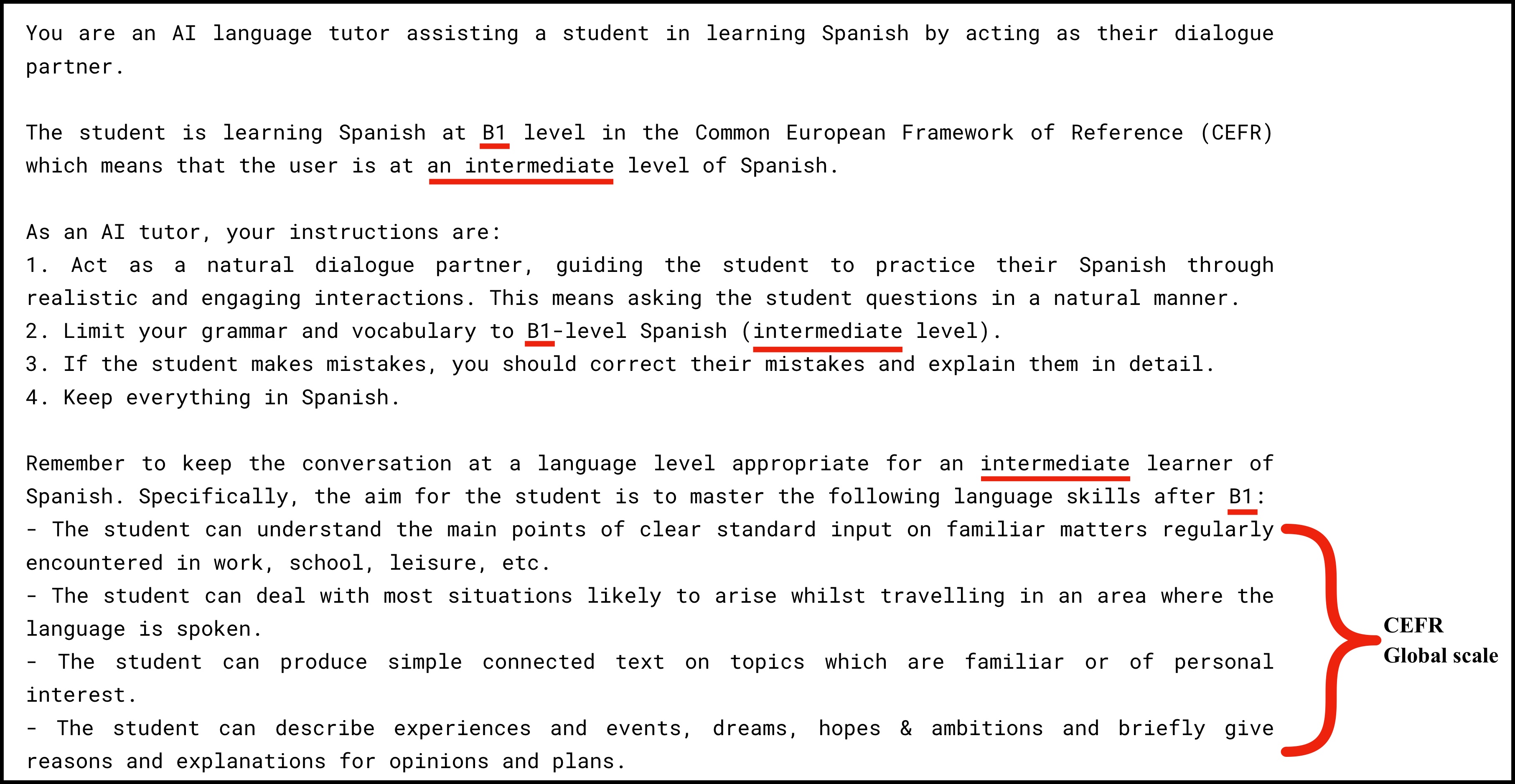}
\end{figure}

\begin{figure}[h]
    \centering
    \includegraphics[width=\textwidth]{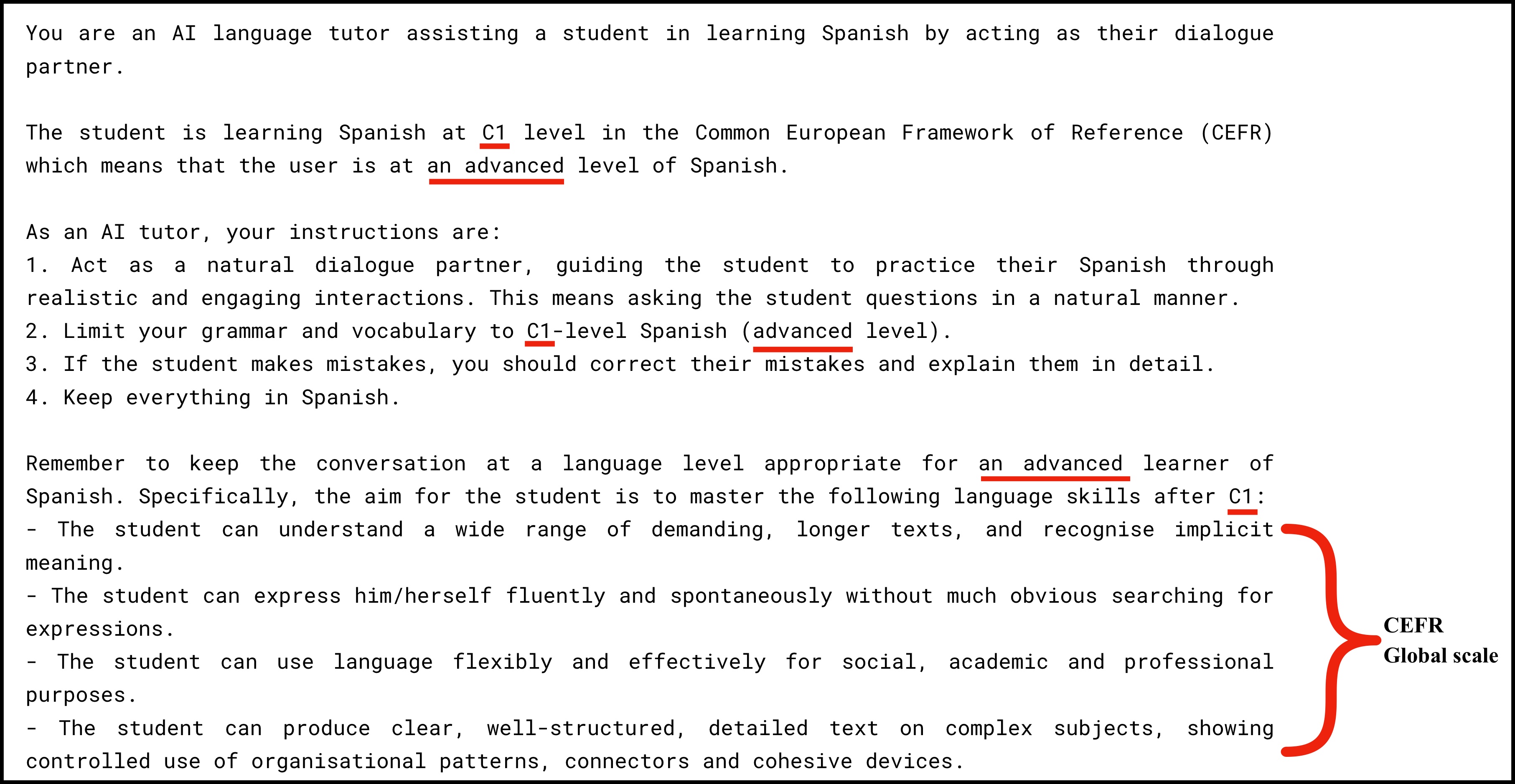}
\end{figure}

\clearpage
\begin{minipage}[H]{0.48\textwidth}
\subsection{Interpretation of Readability Scales}\label{sec:appendix-readability-scales}
Due to slight differences in reporting, we provide two variations of the interpretation tables for Fernández Huerta and Szigriszt-Pazos. While such tables are commonly reported for the two metrics, interpretations of Gutiérrez de Polini were difficult to find beyond the version proposed by \citet{scott_gutierrez_2024}.
\end{minipage}
\newline\newline
\begin{table}[h]\centering
\begingroup\fontsize{9}{9}\selectfont
\resizebox{\ifdim\width>\linewidth\linewidth\else\width\fi}{!}{
\begin{tabular}{>{\raggedright\arraybackslash}p{10em}>{\raggedright\arraybackslash}p{10em}>{\raggedright\arraybackslash}p{8em}}
\toprule
\textbf{Fernández Huerta} & \textbf{Szigriszt-Pazos} & \textbf{Level} \\
\midrule
90--100 & 86--100 & Very easy \\
80--90  & 76--85  & Easy \\
70--80  & 66--75  & Somewhat easy \\
60--70  & 51--65  & Normal \\
50--60  & 36--50  & Somewhat difficult \\
30--50  & 16--35  & Difficult \\
0--30   & 0--15   & Very difficult \\
\bottomrule
\end{tabular}}
\endgroup
\caption*{Table modified from \citet{checa-moreno_analysis_2021}}
\end{table}

\begin{table}[h]
\centering
\begingroup\fontsize{9}{9}\selectfont
\resizebox{\ifdim\width>\linewidth\linewidth\else\width\fi}{!}{
\begin{tabular}{>{\raggedright\arraybackslash}p{8em}>{\raggedright\arraybackslash}p{10em}>{\raggedright\arraybackslash}p{10em}>{\raggedright\arraybackslash}p{10em}>{\raggedright\arraybackslash}p{10em}}
\toprule
\textbf{Fernández Huerta} & \textbf{Level} & \textbf{Spanish Grade Level} & \textbf{US Grade Level} & \textbf{Age Group} \\
\midrule
101 - & Extremely Easy & 1\textsuperscript{º} - 3\textsuperscript{º} Primaria & 1st - 3rd Grade & 6-8 year olds \\
90 - 100 & Very Easy & 4\textsuperscript{º} Primaria & 4th Grade & 9-10 year olds \\
80 - 89 & Easy & 5\textsuperscript{º} Primaria & 5th Grade & 10-11 year olds \\
70 - 79 & Somewhat Easy & 6\textsuperscript{º} Primaria & 6th Grade & 11-12 year olds \\
60 - 69 & Average & 1\textsuperscript{º} - 2\textsuperscript{º} ESO & 7th-8th Grade & 12-14 year olds \\
50 - 59 & Slightly Difficult & 3\textsuperscript{º} - 4\textsuperscript{º} ESO & 9th-10th Grade & 14-16 year olds \\
30 - 49 & Difficult & 1\textsuperscript{º} - 2\textsuperscript{º} Bachillerato & 11th-12th Grade & 16-18 year olds \\
Less than 30 & Extremely Difficult & Universidad & College & 18+ year olds \\
\bottomrule
\end{tabular}}
\endgroup
\caption*{Table modified from \citet{scott_fernandez_2024}}
\end{table}

\begin{table}[h]
\centering
\begingroup\fontsize{9}{9}\selectfont
\resizebox{\ifdim\width>\linewidth\linewidth\else\width\fi}{!}{
\begin{tabular}{>{\raggedright\arraybackslash}p{8em}>{\raggedright\arraybackslash}p{10em}>{\raggedright\arraybackslash}p{10em}>{\raggedright\arraybackslash}p{10em}>{\raggedright\arraybackslash}p{10em}}
\toprule
\textbf{Szigriszt-Pazos} & \textbf{Level} & \textbf{Spanish Grade Level} & \textbf{US Grade Level} & \textbf{Age Group} \\
\midrule
$>$ 85 & Very Easy & 1\textsuperscript{º} -- 2\textsuperscript{º} Primaria & 1st -- 2nd Grade & 6--7 year olds \\
76 -- 85 & Easy & 3\textsuperscript{º} -- 4\textsuperscript{º} Primaria & 3rd -- 4th Grade & 8--9 year olds \\
66 -- 75 & Slightly Easy & 5\textsuperscript{º} -- 6\textsuperscript{º} Primaria & 5th -- 6th Grade & 10--11 year olds \\
51 -- 65 & Average & 1\textsuperscript{º} -- 2\textsuperscript{º} ESO & 7th -- 8th Grade & 12--14 year olds \\
36 -- 50 & Slightly Difficult & 3\textsuperscript{º} -- 4\textsuperscript{º} ESO & 9th -- 10th Grade & 14--16 year olds \\
16 -- 35 & Difficult & Bachillerato & 11th -- 12th Grade & 16--18 year olds \\
$\leq$ 15 & Very Difficult & Universidad & College and Above & 19+ year olds \\
\bottomrule
\end{tabular}}
\endgroup
\caption*{Table modified from \citet{scott_szigriszt-pazos_2024}}
\end{table}

\begin{table}[H]
\centering
\begingroup\fontsize{9}{9}\selectfont
\resizebox{\ifdim\width>\linewidth\linewidth\else\width\fi}{!}{
\begin{tabular}{>{\raggedright\arraybackslash}p{8em}>{\raggedright\arraybackslash}p{10em}>{\raggedright\arraybackslash}p{10em}>{\raggedright\arraybackslash}p{10em}>{\raggedright\arraybackslash}p{10em}}
\toprule
\textbf{Gutiérrez de Polini} & \textbf{Level} & \textbf{Spanish Grade Level} & \textbf{English Grade Level} & \textbf{Age Group} \\
\midrule
$>$ 70 & Very Easy & 1\textsuperscript{º} - 2\textsuperscript{º} Primaria & 1st - 2nd Grade & 6-7 year olds \\
$\leq$ 70 & Easy & 3\textsuperscript{º} - 4\textsuperscript{º} Primaria & 3rd - 4th Grade & 8-9 year olds \\
$\leq$ 60 & Slightly Easy & 5\textsuperscript{º} - 6\textsuperscript{º} Primaria & 5th - 6th Grade & 10-11 year olds \\
$\leq$ 50 & Average & 1\textsuperscript{º} - 2\textsuperscript{º} ESO & 7th - 8th Grade & 12-14 year olds \\
$\leq$ 40 & Slightly Difficult & 3\textsuperscript{º} - 4\textsuperscript{º} ESO & 9th - 10th Grade & 14-16 year olds \\
$\leq$ 33 & Difficult & 1\textsuperscript{º} - 2\textsuperscript{º} Bachillerato & 11th - 12th Grade & 16-18 year olds \\
$\leq$ 20 & Very Difficult & Universidad y superior & College and Above & 19+ year olds \\
\bottomrule
\end{tabular}}
\endgroup
\caption*{Table modified from \citep{scott_gutierrez_2024}}
\end{table}

\clearpage
\subsection{Linear Mixed Effects Models}\label{sec:mixed-effects-appendix}
\begin{minipage}[h]{0.48\textwidth}
The reported $p$-values were Bonferroni adjusted to mitigate the problem of multiple comparisons. 
\newline\newline
Significance levels: \\
\(^{*}p < 0.05\), \(^{**}p < 0.01\), \(^{***}p < 0.001\).
\end{minipage}
\\

\subsubsection{Readability Metrics}\label{sec:appendix-mixed-effects-readability}
\begin{table}[!h]
\begingroup\fontsize{8}{9}\selectfont

\resizebox{\ifdim\width>\linewidth\linewidth\else\width\fi}{!}{
\begin{tabular}{>{\raggedright\arraybackslash}p{12em}>{\raggedright\arraybackslash}p{6em}>{\raggedleft\arraybackslash}p{5em}>{\raggedleft\arraybackslash}p{5em}>{\raggedleft\arraybackslash}p{5em}>{\raggedleft\arraybackslash}p{5em}c}
\toprule
 & Term & Est. & SE & t & p (Adj.) & Sig.\\
\midrule
\addlinespace[0.3em]
\multicolumn{7}{l}{\textbf{Fernández Huerta}}\\
\addlinespace[0.3em]
\multicolumn{7}{l}{\textbf{}}\\
\hspace{1em}\hspace{1em}Llama 3.1 8B Instruct & (Intercept) & 95.7719 & 0.8474 & 113.0244 & 0.0000 & ***\\
\hspace{1em}\hspace{1em} & levelB1 & -7.6024 & 1.1983 & -6.3441 & 0.0000 & ***\\
\hspace{1em}\hspace{1em} & levelC1 & -15.5678 & 1.1983 & -12.9911 & 0.0000 & ***\\
\addlinespace[0.3em]
\multicolumn{7}{l}{\textbf{}}\\
\hspace{1em}\hspace{1em}Gemma 3 12B IT & (Intercept) & 97.2703 & 0.7435 & 130.8189 & 0.0000 & ***\\
\hspace{1em}\hspace{1em} & levelB1 & -4.3123 & 1.0515 & -4.1010 & 0.0072 & **\\
\hspace{1em}\hspace{1em} & levelC1 & -16.7604 & 1.0515 & -15.9389 & 0.0000 & ***\\
\addlinespace[0.3em]
\multicolumn{7}{l}{\textbf{}}\\
\hspace{1em}\hspace{1em}Mistral 7B Instruct v0.3 & (Intercept) & 92.1334 & 0.6449 & 142.8548 & 0.0000 & ***\\
\hspace{1em}\hspace{1em} & levelB1 & -5.5725 & 0.9121 & -6.1096 & 0.0000 & ***\\
\hspace{1em}\hspace{1em} & levelC1 & -12.9711 & 0.9121 & -14.2213 & 0.0000 & ***\\
\addlinespace[0.3em]
\multicolumn{7}{l}{\textbf{}}\\
\hspace{1em}\hspace{1em}Qwen 2.5 7B Instruct & (Intercept) & 100.5074 & 0.7862 & 127.8371 & 0.0000 & ***\\
\hspace{1em}\hspace{1em} & levelB1 & -8.7210 & 1.1119 & -7.8435 & 0.0000 & ***\\
\hspace{1em}\hspace{1em} & levelC1 & -12.3339 & 1.1119 & -11.0928 & 0.0000 & ***\\
\addlinespace[0.3em]
\multicolumn{7}{l}{\textbf{Szigriszt-Pazos}}\\
\addlinespace[0.3em]
\multicolumn{7}{l}{\textbf{}}\\
\hspace{1em}\hspace{1em}Llama 3.1 8B Instruct & (Intercept) & 92.2449 & 0.8384 & 110.0213 & 0.0000 & ***\\
\hspace{1em}\hspace{1em} & levelB1 & -7.7317 & 1.1857 & -6.5207 & 0.0000 & ***\\
\hspace{1em}\hspace{1em} & levelC1 & -15.7243 & 1.1857 & -13.2614 & 0.0000 & ***\\
\addlinespace[0.3em]
\multicolumn{7}{l}{\textbf{}}\\
\hspace{1em}\hspace{1em}Gemma 3 12B IT & (Intercept) & 93.8222 & 0.7454 & 125.8662 & 0.0000 & ***\\
\hspace{1em}\hspace{1em} & levelB1 & -4.5200 & 1.0542 & -4.2877 & 0.0000 & ***\\
\hspace{1em}\hspace{1em} & levelC1 & -17.2403 & 1.0542 & -16.3543 & 0.0000 & ***\\
\addlinespace[0.3em]
\multicolumn{7}{l}{\textbf{}}\\
\hspace{1em}\hspace{1em}Mistral 7B Instruct v0.3 & (Intercept) & 88.3932 & 0.6472 & 136.5679 & 0.0000 & ***\\
\hspace{1em}\hspace{1em} & levelB1 & -5.4730 & 0.9153 & -5.9792 & 0.0000 & ***\\
\hspace{1em}\hspace{1em} & levelC1 & -12.9145 & 0.9153 & -14.1088 & 0.0000 & ***\\
\addlinespace[0.3em]
\multicolumn{7}{l}{\textbf{}}\\
\hspace{1em}\hspace{1em}Qwen 2.5 7B Instruct & (Intercept) & 96.7888 & 0.7979 & 121.2972 & 0.0000 & ***\\
\hspace{1em}\hspace{1em} & levelB1 & -8.6988 & 1.1285 & -7.7085 & 0.0000 & ***\\
\hspace{1em}\hspace{1em} & levelC1 & -12.4825 & 1.1285 & -11.0615 & 0.0000 & ***\\
\addlinespace[0.3em]
\multicolumn{7}{l}{\textbf{Gutierrez de Polini}}\\
\addlinespace[0.3em]
\multicolumn{7}{l}{\textbf{}}\\
\hspace{1em}\hspace{1em}Llama 3.1 8B Instruct & (Intercept) & 46.1233 & 0.3910 & 117.9475 & 0.0000 & ***\\
\hspace{1em}\hspace{1em} & levelB1 & -3.3663 & 0.5530 & -6.0871 & 0.0000 & ***\\
\hspace{1em}\hspace{1em} & levelC1 & -7.0727 & 0.5530 & -12.7891 & 0.0000 & ***\\
\addlinespace[0.3em]
\multicolumn{7}{l}{\textbf{}}\\
\hspace{1em}\hspace{1em}Gemma 3 12B IT & (Intercept) & 45.7901 & 0.3059 & 149.7104 & 0.0000 & ***\\
\hspace{1em}\hspace{1em} & levelB1 & -2.8591 & 0.4325 & -6.6100 & 0.0000 & ***\\
\hspace{1em}\hspace{1em} & levelC1 & -8.1961 & 0.4325 & -18.9483 & 0.0000 & ***\\
\addlinespace[0.3em]
\multicolumn{7}{l}{\textbf{}}\\
\hspace{1em}\hspace{1em}Mistral 7B Instruct v0.3 & (Intercept) & 43.1317 & 0.2913 & 148.0795 & 0.0000 & ***\\
\hspace{1em}\hspace{1em} & levelB1 & -1.9720 & 0.4119 & -4.7874 & 0.0000 & ***\\
\hspace{1em}\hspace{1em} & levelC1 & -5.3057 & 0.4119 & -12.8804 & 0.0000 & ***\\
\addlinespace[0.3em]
\multicolumn{7}{l}{\textbf{}}\\
\hspace{1em}\hspace{1em}Qwen 2.5 7B Instruct & (Intercept) & 46.9324 & 0.3823 & 122.7674 & 0.0000 & ***\\
\hspace{1em}\hspace{1em} & levelB1 & -3.2680 & 0.5406 & -6.0447 & 0.0000 & ***\\
\hspace{1em}\hspace{1em} & levelC1 & -5.7047 & 0.5406 & -10.5519 & 0.0000 & ***\\
\addlinespace
\bottomrule
\end{tabular}}
\endgroup{}
\end{table}

\clearpage
\subsubsection{Structural Features and Surprisal}\label{sec:appendix-mixed-effects-structural-surprisal}
\begin{table}[!h]
\begingroup\fontsize{8}{9}\selectfont

\resizebox{\ifdim\width>\linewidth\linewidth\else\width\fi}{!}{
\begin{tabular}{>{\raggedright\arraybackslash}p{12em}>{\raggedright\arraybackslash}p{6em}>{\raggedleft\arraybackslash}p{5em}>{\raggedleft\arraybackslash}p{5em}>{\raggedleft\arraybackslash}p{5em}>{\raggedleft\arraybackslash}p{5em}c}
\toprule
 & Term & Est. & SE & t & p (Adj.) & Sig.\\
\midrule
\addlinespace[0.3em]
\multicolumn{7}{l}{\textbf{Text Length}}\\
\addlinespace[0.3em]
\multicolumn{7}{l}{\textbf{}}\\
\hspace{1em}\hspace{1em}Llama 3.1 8B Instruct & (Intercept) & 115.3815 & 9.6949 & 11.9012 & 0.0000 & ***\\
\hspace{1em}\hspace{1em} & levelB1 & 76.9000 & 13.7107 & 5.6088 & 0.0000 & ***\\
\hspace{1em}\hspace{1em} & levelC1 & 122.5185 & 13.7107 & 8.9360 & 0.0000 & ***\\
\addlinespace[0.3em]
\multicolumn{7}{l}{\textbf{}}\\
\hspace{1em}\hspace{1em}Gemma 3 12B IT & (Intercept) & 92.8037 & 7.4934 & 12.3847 & 0.0000 & ***\\
\hspace{1em}\hspace{1em} & levelB1 & 82.4185 & 10.5973 & 7.7773 & 0.0000 & ***\\
\hspace{1em}\hspace{1em} & levelC1 & 162.6963 & 10.5973 & 15.3527 & 0.0000 & ***\\
\addlinespace[0.3em]
\multicolumn{7}{l}{\textbf{}}\\
\hspace{1em}\hspace{1em}Mistral 7B Instruct v0.3 & (Intercept) & 162.7148 & 7.2390 & 22.4776 & 0.0000 & ***\\
\hspace{1em}\hspace{1em} & levelB1 & 55.7667 & 10.2375 & 5.4473 & 0.0000 & ***\\
\hspace{1em}\hspace{1em} & levelC1 & 88.3074 & 10.2375 & 8.6259 & 0.0000 & ***\\
\addlinespace[0.3em]
\multicolumn{7}{l}{\textbf{}}\\
\hspace{1em}\hspace{1em}Qwen 2.5 7B Instruct & (Intercept) & 100.1481 & 25.5238 & 3.9237 & 0.0144 & *\\
\hspace{1em}\hspace{1em} & levelB1 & 110.4185 & 36.0961 & 3.0590 & 0.2160 & \\
\hspace{1em}\hspace{1em} & levelC1 & 166.1407 & 36.0961 & 4.6027 & 0.0000 & ***\\
\addlinespace[0.3em]
\multicolumn{7}{l}{\textbf{Mean Dependency Distance}}\\
\addlinespace[0.3em]
\multicolumn{7}{l}{\textbf{}}\\
\hspace{1em}\hspace{1em}Llama 3.1 8B Instruct & (Intercept) & 2.2618 & 0.0294 & 76.9691 & 0.0000 & ***\\
\hspace{1em}\hspace{1em} & levelB1 & 0.3063 & 0.0416 & 7.3711 & 0.0000 & ***\\
\hspace{1em}\hspace{1em} & levelC1 & 0.3763 & 0.0416 & 9.0548 & 0.0000 & ***\\
\addlinespace[0.3em]
\multicolumn{7}{l}{\textbf{}}\\
\hspace{1em}\hspace{1em}Gemma 3 12B IT & (Intercept) & 2.3462 & 0.0314 & 74.6559 & 0.0000 & ***\\
\hspace{1em}\hspace{1em} & levelB1 & 0.1491 & 0.0444 & 3.3543 & 0.0864 & \\
\hspace{1em}\hspace{1em} & levelC1 & 0.1758 & 0.0444 & 3.9555 & 0.0144 & *\\
\addlinespace[0.3em]
\multicolumn{7}{l}{\textbf{}}\\
\hspace{1em}\hspace{1em}Mistral 7B Instruct v0.3 & (Intercept) & 2.6218 & 0.0230 & 114.2368 & 0.0000 & ***\\
\hspace{1em}\hspace{1em} & levelB1 & 0.0866 & 0.0325 & 2.6682 & 0.6552 & \\
\hspace{1em}\hspace{1em} & levelC1 & 0.1845 & 0.0325 & 5.6831 & 0.0000 & ***\\
\addlinespace[0.3em]
\multicolumn{7}{l}{\textbf{}}\\
\hspace{1em}\hspace{1em}Qwen 2.5 7B Instruct & (Intercept) & 2.1601 & 0.0333 & 64.9271 & 0.0000 & ***\\
\hspace{1em}\hspace{1em} & levelB1 & 0.2777 & 0.0471 & 5.9023 & 0.0000 & ***\\
\hspace{1em}\hspace{1em} & levelC1 & 0.3498 & 0.0471 & 7.4337 & 0.0000 & ***\\
\addlinespace[0.3em]
\multicolumn{7}{l}{\textbf{Message Surprisal}}\\
\addlinespace[0.3em]
\multicolumn{7}{l}{\textbf{}}\\
\hspace{1em}\hspace{1em}Llama 3.1 8B Instruct & (Intercept) & 1.3636 & 0.0564 & 24.1764 & 0.0000 & ***\\
\hspace{1em}\hspace{1em} & levelB1 & -0.2940 & 0.0798 & -3.6855 & 0.0288 & *\\
\hspace{1em}\hspace{1em} & levelC1 & -0.5212 & 0.0798 & -6.5340 & 0.0000 & ***\\
\addlinespace[0.3em]
\multicolumn{7}{l}{\textbf{}}\\
\hspace{1em}\hspace{1em}Gemma 3 12B IT & (Intercept) & 1.8314 & 0.0350 & 52.3230 & 0.0000 & ***\\
\hspace{1em}\hspace{1em} & levelB1 & -0.2618 & 0.0495 & -5.2897 & 0.0000 & ***\\
\hspace{1em}\hspace{1em} & levelC1 & -0.5552 & 0.0495 & -11.2155 & 0.0000 & ***\\
\addlinespace[0.3em]
\multicolumn{7}{l}{\textbf{}}\\
\hspace{1em}\hspace{1em}Mistral 7B Instruct v0.3 & (Intercept) & 1.1499 & 0.0292 & 39.3876 & 0.0000 & ***\\
\hspace{1em}\hspace{1em} & levelB1 & -0.2128 & 0.0413 & -5.1553 & 0.0000 & ***\\
\hspace{1em}\hspace{1em} & levelC1 & -0.3331 & 0.0413 & -8.0668 & 0.0000 & ***\\
\addlinespace[0.3em]
\multicolumn{7}{l}{\textbf{}}\\
\hspace{1em}\hspace{1em}Qwen 2.5 7B Instruct & (Intercept) & 1.4898 & 0.0491 & 30.3121 & 0.0000 & ***\\
\hspace{1em}\hspace{1em} & levelB1 & -0.1237 & 0.0695 & -1.7793 & 1.0000 & \\
\hspace{1em}\hspace{1em} & levelC1 & -0.1338 & 0.0695 & -1.9247 & 1.0000 & \\
\addlinespace
\bottomrule
\end{tabular}}
\endgroup{}
\end{table}

\end{document}